\pdfoutput=1

\documentclass[11pt]{article}

\usepackage{acl}

\usepackage{times}
\usepackage{bm}
\usepackage{latexsym}
\usepackage{amsmath}
\usepackage{amssymb}
\usepackage{bbm}
\usepackage{cleveref}
\usepackage{float}
\usepackage{fix-cm}
\usepackage{pifont}

\definecolor{trend}{HTML}{245D95}

\usepackage[T1]{fontenc}

\usepackage[utf8]{inputenc}

\usepackage{microtype}

\usepackage{inconsolata}

\usepackage{graphicx}

\newcounter{manualhypo}
\Crefname{manualhypo}{Hypothesis}{Hypotheses}

\usepackage[most]{tcolorbox}
\usepackage{cleveref}
\usepackage{xcolor}

\newenvironment{ManualHypo}[2]{%
  \renewcommand{\themanualhypo}{#2}
  \refstepcounter{manualhypo}
  \begin{tcolorbox}[
  enhanced,
  colback=gray!10,
  colframe=black!50,
  boxrule=0.4pt,
  arc=4pt,
  left=6pt, right=6pt, top=4pt, bottom=4pt,
  fonttitle=\bfseries,
  title=Hypothesis~#2,
  ]
}{%
  \end{tcolorbox}
}

\title{Language Models Grow Less Humanlike beyond Phase Transition}

\author{Tatsuya Aoyama \hspace{10pt}
  Ethan Gotlieb Wilcox \\
  Department of Linguistics, Georgetown University \\
  \{\emldisplay{ta571@georgetown.edu}{ta571}, \emldisplay{ethan.wilcox@georgetown.edu}{ethan.wilcox}\}\texttt{@georgetown.edu}}

\begin{document}

\maketitle

\begin{abstract}

LMs' alignment with human reading behavior (i.e. \textit{psychometric predictive power; PPP}) is known to improve during pretraining up to a tipping point, beyond which it either plateaus or degrades. Various factors, such as word frequency, recency bias in attention, and context size, have been theorized to affect PPP, yet there is no current account that explains \emph{why} such a tipping point exists, and how it interacts with LMs' pretraining dynamics more generally. We hypothesize that the underlying factor is a pretraining \textit{phase transition}, characterized by the rapid emergence of specialized attention heads. We conduct a series of correlational and causal experiments to show that such a phase transition is responsible for the tipping point in PPP. We then show that, rather than producing attention patterns that contribute to the degradation in PPP, phase transitions alter the subsequent learning dynamics of the model, such that further training keeps damaging PPP.
\end{abstract}

\section{Introduction}\label{sec:intro}

The rise of neural-network-based language models (LMs) that can produce fluent, humanlike linguistic outputs has led to their increased use as cognitive models of human language processing \citep{warstadt2024artificial}.
In particular, an active area of research has studied how well LMs are aligned with human incremental processing behaviors. This is often measured by how well LMs' output probabilities predict various human reading time metrics, and is referred to as a model's \textit{psychometric predictive power} or \textit{PPP}. PPP
has been widely studied for English (\citealp[e.g., ][\textit{inter alia}]{wilcox-etal-2020-predictive, shain-etal-2024-large, oh-etal-2022-gpt, oh-schuler-2023-surprisal}), Japanese \cite{kuribayashi-etal-2021-lower}, multilingually \cite{wilcox-etal-2023-testing}, and for nonnative Englishes \cite{aoyama-schneider-2024-modeling}, where it has been found that LMs' outputs are robustly correlated with human reading times.

\begin{figure}[t]
    \hspace{-20pt}
    \input{latex/figures/splashy.pgf}
    \vspace{-20pt}
    \caption{Phase transition ($x$-axis) and PPP peaks ($y$-axis) closely coincide with each other across the corpora studied in this work, except for MECO. Numbers in the parentheses are batch sizes; for Pythia models, regardless of the model size, the batch size is 1,024. \textbf{\textcolor{orange}{Orange}} dashed lines, \textbf{\textcolor{trend}{blue}} solid lines, and \textbf{\textcolor{gray}{gray}} dashed lines represent the \textbf{\textcolor{orange}{phase transition hypothesis}}, \textbf{\textcolor{gray}{2 billion tokens hypothesis}}, and the \textbf{\textcolor{trend}{observed trendline}} drawn from the data, respectively. We find that the experimental results closely align with our phase transition hypothesis, meaning that the PPP peaks correspond to LM phase transition, beyond which PPP starts degrading. *$p <.05$; **$p < .01$.}
    \vspace{-10pt}
    \label{fig:splashy}
\end{figure}

One outstanding puzzle in these data is the observation that transformer-based LMs become maximally aligned with human sentence processing relatively early in pretraining, after which their fit to human data either plateaus or decreases. We will refer to the maxima of psychological fit during training as model's \define{tipping point}. Such tipping points present a conundrum: Why does the fit to human data decrease, even as the model's language modeling loss goes down? And what is the tipping point's underlying cause?

Previous work has observed that tipping points in models tend to occur after around 2 billion words of pretraining, and has suggested that this number---2 billion---is operative \citep{oh-schuler-2023-transformer}. We refer to this as the \define{2 Billion Hypothesis}. In contrast, we suspect that the tipping point at 2 billion tokens is coincidental rather than causal. Instead, we hypothesize that the reversal of PPP is related to more general pretraining dynamics, specifically the presence of \define{phase transitions}, or periods in pretraining when new model capabilities emerge rapidly. While phase transition can be a general term used to describe any rapid change in model behaviors, we focus on what we term \define{specialized heads phrase transitions}, characterized by the emergence of specialized attention patterns relatively early in pretraining \citep{olsson2022context, chen2024sudden}. For brevity, we refer to these simply as ``phase transition'' (see \Cref{sec:relevant:pt} for a more precise definition). We dub this hypothesis the \textbf{Phase Transition Hypothesis} (see \Cref{hypo:1}) and conduct a series of experiments, both correlational and causal, to test it (\Cref{fig:splashy}).

Our key findings include: (1) the tipping points of PPP are strongly correlated with phase transition (\Cref{sec:exp1}); (2) a series of ablation experiments show some attention heads that formed during phase transition have some effect on PPP, but the overall results are mixed and the PPP degradation cannot be reliably attributed to specific circuits that form during the phase transition (negates \Cref{hypo:2a}; see \Cref{sec:exp2}); and (3) regularization during pretraining delays the phase transition, also resulting in a delayed PPP tipping point, indicating that phase transition alters the learning dynamics, such that further training results in PPP degradation (supports \Cref{hypo:2b}; see \Cref{sec:exp3}).\footnote{Code and data available at \url{https://github.com/picol-georgetown/pt-and-ppp}.}

\section{Relevant Work}\label{sec:relevant}
\subsection{Sentence Processing}\label{sec:relevant:sp}

One popular theory to explain human sentence processing is \emph{surprisal theory}. Surprisal theory posits that the processing difficulty of a word is proportional to its information content, quantified as its surprisal, or in-context, negative log probability \cite{LEVY20081126, hale-2001-probabilistic}. Surprisal theory is supported by numerous studies showing a tight linear relationship between incremental processing times and surprisal, across various datasets \citep{SMITH2013302, shain-etal-2024-large} and languages (\citealp{wilcox-etal-2023-testing}).\footnote{Although cf. \citealp{meister-etal-2021-revisiting} who find a slightly \emph{super}-linear relationship.}

Rather than using LMs as a gold proxy for word predictability, other studies have compared how surprisals obtained from various LMs show different fits to human reading data. \citet{goodkind-bicknell-2018-predictive} find that LM quality, as measured in perplexity, linearly correlates with PPP, such that surprisals obtained from better LMs are more predictive of human reading time. This relationship between LM quality and PPP has been dubbed the \textit{quality-power hypothesis} and replicated with different sets of LMs \cite{wilcox-etal-2020-predictive, oh-etal-2022-gpt, oh-schuler-2023-surprisal}, and cross-lingually \cite{wilcox-etal-2023-language}.

Instead of comparing fully trained LMs, \citet{oh-schuler-2023-transformer} find that the quality-power relationship changes during pretraining, which motivates the current study. Specifically, they find that PPP keeps improving until around 2 billion tokens of pretraining (for all Pythia variants) and then starts degrading beyond that point, although perplexity keeps improving. In other words, the quality-power correlation is \textit{positive} (better the model, higher the PPP) until a certain point and \textit{negative} (better the model, lower the PPP) after it. We call this point a \define{tipping point} in this paper, and define it as the maximum PPP obtained during pretraining. In fact, \citet{kuribayashi-etal-2021-lower} seems to be the first to report this trend, where they find similar tipping points for English and Japanese. This tipping point has been questioned in \citet{aoyama-schneider-2024-modeling}, where they observe a \textit{negative} quality-power correlation at orders of magnitude smaller pretraining amounts for crosslingual LMs.

\subsection{Factors Affecting LMs' PPP}\label{sec:relevant:factors}

Besides the amount of pretraining, several other factors have been found to affect LMs' PPP.\footnote{For a more extensive list, we refer the readers to Table 1 of \citealp{kuribayashi-etal-2022-context}} One key factor is model size, where larger models tend to perform worse in modeling human sentence processing compared to smaller models. \citet{oh-etal-2022-gpt} found that structural parsers and $n$-gram models often matched or outperformed GPT2 models, with the smallest GPT2 variant yielding the best results. \citet{oh-schuler-2023-surprisal} tested additional GPT and OPT variants confirming this trend. Another crucial factor is context size, as limiting the inference-time context window has been shown to improve PPP. \citet{kuribayashi-etal-2022-context} observed that shorter contexts, particularly bigrams, resulted in better PPP, especially for Japanese. Additionally, introducing recency bias in attention mechanisms improves PPP \cite{de-varda-marelli-2024-locally}, particularly when applied during both training and inference \cite{clark-etal-2024-linear}.

Lexical factors also play a role in PPP degradation among larger models. \citet{oh-schuler-2023-surprisal} found that larger models tend to assign lower surprisals to open-class words (e.g., nouns and adjectives) and named entities, deviating more from human reading patterns. \citet{oh-etal-2024-frequency} further linked this effect to token frequency, showing that larger models exhibit particularly low alignment with human reading time for infrequent tokens, especially beyond a certain point in pretraining (2B tokens). These studies suggest that model size, long context utilization, and pretraining dynamics interact with each other to produce PPP degradation. We propose that one way to tie these pieces together is by finding a common cause, which we hypothesize to be phase transitions.

\subsection{Phase Transition}\label{sec:relevant:pt}

What exactly is a phase transition, and how is it measured? Abrupt changes in model behaviors, which cannot be predicted by a scaling law \cite{kaplan2020scalinglawsneurallanguage}\footnote{See \citealp{caballero2023broken} for a scaling law with `breaks'.}, have been studied widely (e.g., \citealp{wei2022emergent, srivastava2023beyond}). In this paper, we specifically focus on what we call \define{specialized heads phase transition}, which is characterized by the rapid emergence of specialized attention patterns \cite{elhage2021mathematical, olsson2022context, chen2024sudden}, relatively early in pretraining. For brevity, we use the term ``phase transition.''

We operationalize the phase transition using two well-studied phenomena, the presence of induction heads \cite{elhage2021mathematical, olsson2022context} and syntactic attention structure (SAS; \citealp{chen2024sudden}). \citet{elhage2021mathematical, olsson2022context}
find that a specialized head called an induction head emerges at a certain point during pretraining, and that this head is characterized by its distinctive copying behavior, where the head attends to the previous occurrence of the same or similar bigram within the context. We introduce the metric used to detect induction heads in \Cref{eq:pms}. \citet{olsson2022context} further claim that the emergence of this induction head is primarily responsible for most of the in-context learning (ICL) abilities (see \Cref{eq:icl-score}).

Similarly, \citet{chen2024sudden} find that LMs demonstrate a sudden emergence of attention patterns that mirror syntactic dependency edges, which they call SAS. Similar to induction heads, the emergence of SAS is followed by an abrupt improvement in syntactic abilities as measured by performance on a syntactic probing benchmark, BLiMP \cite{warstadt-etal-2020-blimp-benchmark}.

\subsection{Research Questions}\label{sec:rq}
In light of this, we ask the following research question: \textbf{to what extent is phase transition responsible for the degradation in the alignment between human and LM sentence processing?} We present three hypotheses, outlined below:

\begin{ManualHypo}{hypo:1}{1}
\label{hypo:1}
\textbf{Phase Transition Hypothesis:} PPP degradation coincides with phase transition.
\end{ManualHypo}

\noindent If this is the case, then we need to account for \textit{how} and \textit{why} phase transition causes the PPP degradation, for which we have two hypotheses:

\begin{ManualHypo}{hypo:2a}{2a}
\label{hypo:2a}
\textbf{Persistence Hypothesis:} Circuits that emerge during phase transition cause PPP degradation.
\end{ManualHypo}

\noindent This hypothesis predicts that certain circuits emerge at the phase transition, persist afterward, and contribute to the degradation in PPP. Alternatively, we may hypothesize that the degradation arises from pretraining dynamics \textit{after} the phase transition. This view predicts that preventing PPP degradation requires suppressing the phase transition itself, rather than deactivating specific circuits, leading to our final hypothesis:
\begin{ManualHypo}{hypo:2b}{2b}
\label{hypo:2b}
\textbf{Dynamics Hypothesis:} Altered training dynamics causes PPP degradation.
\end{ManualHypo}
\noindent 
Crucially, \Cref{hypo:1} serves as a prerequisite for the others, while \Cref{hypo:2a} and \Cref{hypo:2b} may be either competing or complementary. Put differently, \Cref{hypo:2a} attributes PPP degradation to circuits that emerge \textit{during} the phase transition, whereas \Cref{hypo:2b} attributes it to circuits that emerge \textit{after} the phase transition.

\section{Methods}\label{sec:method}

\subsection{Models and Checkpoints}\label{sec:method:models}

Because online sentence processing is a predominantly left-to-right process, we only consider autoregressive LMs, a common practice adopted by many others (see \citealp{meister-etal-2023-locally} for a rare exception). Since training LLMs is expensive, we primarily focus on the Pythia family \cite{biderman2023pythiasuiteanalyzinglarge}. To our knowledge, Pythia is a rare, if not the only, LM family that satisfies the following conditions necessary in this paper: (1) open-sourced, (2) available in different sizes, and importantly, (3) various checkpoints, especially early in pretraining, are available. Bloom \cite{bigscience_workshop_2022} is another candidate; however, because the available checkpoints are likely after the phase transition, we did not include it in this study.

We also train several GPT2 models \cite{radford2019language} and save log-spaced checkpoints at $\{500K, 1M, 2M, ..., 256M\}$ tokens for the first 10 checkpoints, and evenly-spaced checkpoints for the last 20 checkpoints at $\{0.5B, 1B, 1.5B ..., 10B\}$ tokens. This is a superset of the available Pythia checkpoints for the first 10B tokens of pretraining, which allows for direct comparisons among different model sizes and families at the same pretraining amounts. All custom-trained GPT2 models have 2 layers, 8 heads (per layer), a hidden dimension of 768, and a total parameter count of 53M. See \Cref{tab-hyperparams} in \Cref{sec:appendix-hyperparams} for the list of architecture and training hyperparameters. Each run took $\approx$70 hours on a single A6000.

\subsection{Data}\label{sec:method:data}
For custom trained models, we use a sample of 1B tokens from the English subcorpus of \texttt{CC100} (\citealt{conneau-etal-2020-unsupervised, wenzek-etal-2020-ccnet}; Apache License 2.0). For reading time data, we use 3 eye-tracking corpora: Dundee \cite{kennedy2003dundee}, Provo (\citealp{luke2018provo}; CC-BY 4.0)\footnote{\url{https://osf.io/sjefs/}}, and MECO \cite{siegelman2022expanding} release 1.2.\footnote{\url{https://osf.io/3527a/}}, and 1 self-paced reading time corpus: Natural Stories (\citealp{futrell2021natural}; CC BY-NC-SA). For eye-tracking corpora, we focus on the gaze duration.\footnote{See \citet{shain-etal-2024-large} for comparisons across different eye-tracking metrics, which yield similar results.)}

\subsection{Calculating PPP}

Following previous work (e.g. \citealp{goodkind-bicknell-2018-predictive, wilcox-etal-2020-predictive, wilcox-etal-2023-language}, \textit{inter alia}), we operationalize PPP as the difference in log-likelihood (LL) between two linear models (delta log-likelihood; \dll):
\begin{equation}
    \dll = \text{LL}_{f_{\text{base+surp}}}-\text{LL}_{f_{\text{base}}}
\end{equation}
\noindent where $f_{\text{base+surp}}$ and $f_{\text{base}}$ are linear models that predict human reading times using baseline features with and without LM surprisals, respectively. Finally, surprisal of a word $w_i$ in context is its negative log probability: $-\log \text{P}(w_i|\mathbf{w}_{<i})$. See \Cref{appendix:ppp} for the full list of features included in the regression models.

\section{Experiment 1: PPP Peaks at Phase Transition}\label{sec:exp1}

\subsection{Method}\label{sec:exp1:method}

As we want to show that PPP peaks at the phase transition, defined by the emergence of SAS and induction heads, we outline the definition of the metrics we use to characterize them below.

\noindent\textbf{Unlabeled Attachment Score (UAS)}. UAS is a commonly used metric for dependency parsing, and in the context of SAS, conceptually, it measures how well a given LM's attention pattern matches the dependency edges between words, ignoring the relation labels. In other words, UAS is the proportion of words, such that the highest attention weight lies between the word and its parent for the best-performing head for the given relation type. See \Cref{appendix:uas} for details.

\noindent\textbf{Prefix-matching Score (PS).} Induction heads are defined by the copying behavior, such that if the model has seen an \ab sequence and the current token is \toka, then an induction head is a head that promotes the prediction of \tokb as the next token, completing the \abab sequence. 
We quantify this behavior using PS \cite{olsson2022context}. The intuition behind this metric is to measure, when predicting what follows the second occurrence of $A$, how much a model is referencing what followed its first occurrence.

Given a random sequence of tokens $\mathbf{x}$ repeated twice, PS of a head $h$ at layer $l$ is its average attention from the source token $x_i$ to the next token of its previous occurrence:
\begin{equation}\label{eq:pms}
\text{PS}=\frac{1}{|\mathbf{x}|-1}\sum_{i=|\mathbf{x}|+1}^{2|\mathbf{x}|}\mathrm{\alpha}^{(h,l)}(x_i, x_{i-(|\mathbf{x}|-1)})
\end{equation}
In our experiments, we use a random sequence of length 50, following the \texttt{TransformerLens} library \cite{nanda2022transformerlens}, so $|\mathbf{x}|=50$. Note that $\mathbf{x}$ is a completely random sequence of token IDs; hence, attention patterns are not an artifact of data contamination or memorization.

\noindent\textbf{Breakthrough Points.} \citet{chen2024sudden} defines \textit{breakthrough} in some metric $f$ as the acceleration point, which maximizes the growth in slope, measured in some discrete intervals $\Delta$. To make it less susceptible to surface fluctuations, we define them as the first checkpoint $c$ at which a certain metric $f$ exceeds a given threshold $t$: $\min \{c \in \mathcal{C}|f(c)>t\}$, where $\mathcal{C}$ is the predefined set of checkpoints described in \Cref{sec:method:models}. In our experiments, we use $t=0.1$ for both UAS and PS. We validate these implementations of breakthroughs as well as the measurements of UAS and PS by showing that the sudden rises in UAS and PS are followed by the improvement in BLiMP and ICL scores, respectively (see \Cref{appendix:pt-details} for these results).

\subsection{Results}\label{sec:exp1:result}

\Cref{fig:splashy} shows on the $x$-axis the number of pretraining tokens at which phase transition occurred, as defined by UAS (top) and by PS (bottom), and on the $y$-axis the number of pretraining tokens at which the peak \dll was observed. The \textcolor{gray}{gray lines} represent the \textcolor{gray}{\define{2 Billion Hypothesis}} based on \citet{oh-schuler-2023-transformer}, where transformer models are predicted to reach the highest \dll at around 2 billion pretraining tokens. The \textcolor{orange}{orange lines} represent our \textcolor{orange}{\define{Phase Transition Hypothesis}}, where the peak \dll is predicted to happen at around the same time as the phase transition. The \textcolor{trend}{blue lines} are the \textbf{\textcolor{trend}{observed trendlines}}, and we can see that they closely match the Phase Transition Hypothesis. This trend holds across 3 orders of magnitude, meaning that the \dll peaks occur at vastly different times across different models, yet they all co-occur with their respective phase transition points, although this trend was not found in MECO (see \Cref{sec:limitation} for known issues of MECO). Note that we included Pythia models used in \citet{oh-schuler-2023-transformer}, and GPT2 models trained from scratch with various batch sizes, as we find that the phase transition point is a function of not only the number of pretraining tokens, but also the number of updates (hence changing batch size affects phase transition points). We leave a precise characterization of phase transition points to future work.
For a complete visualization of the full \dll trajectory, see \Cref{fig:coinciding} (\Cref{appendix:full-results}), which also includes the 0- and 1-layer models. These are predicted to exhibit qualitatively different behaviors, with the 0-layer model indeed showing virtually no PPP degradation.

The concurrence of PPP peaks and phase transitions seems robust; however, these observations are still correlational, and we need more evidence to make a causal claim. As such, we consider two potential hypotheses, outlined earlier: (1) phase transition develops some circuits that hurt PPP (Persistence Hypothesis), and/or (2) phase transition changes the course of pretraining dynamics, in a way such that further training hurts PPP (Dynamics Hypothesis).
We test each hypothesis in the following experiments.

\section{Experiment 2: PPP Degradation Cannot be Attributed to Specific Heads}\label{sec:exp2}

If phase transition creates certain circuits responsible for the lower PPP (\Cref{hypo:2a}; Persistence Hypothesis), ablating those circuits should improve PPP for models that have undergone the phase transition. In this experiment, we ablate each head one at a time and investigate if the ablation of specialized heads that form during phase transition (i.e. SAS and induction heads) improves PPP more than other heads.
\begin{table*}[t]
    \centering
    \footnotesize
    \setlength{\tabcolsep}{.12em}
\begin{tabular}{c|rrrr|rrrr|c|rrrr|rrrr|rrrr}
\toprule
 &\multicolumn{4}{c|}{GPT2 (1 layer)}&\multicolumn{4}{c|}{GPT2 (2 layers)}& &\multicolumn{4}{c|}{Pythia-70M}&\multicolumn{4}{c|}{Pythia-160M}&\multicolumn{4}{c}{Pythia-410M}\\
\midrule
 &\multicolumn{1}{c}{DU}&\multicolumn{1}{c}{ME}&\multicolumn{1}{c}{PR}&\multicolumn{1}{c|}{NS}&\multicolumn{1}{c}{DU}&\multicolumn{1}{c}{ME}&\multicolumn{1}{c}{PR}&\multicolumn{1}{c|}{NS}& &\multicolumn{1}{c}{DU}&\multicolumn{1}{c}{ME}&\multicolumn{1}{c}{PR}&\multicolumn{1}{c|}{NS}&\multicolumn{1}{c}{DU}&\multicolumn{1}{c}{ME}&\multicolumn{1}{c}{PR}&\multicolumn{1}{c|}{NS}&\multicolumn{1}{c}{DU}&\multicolumn{1}{c}{ME}&\multicolumn{1}{c}{PR}&\multicolumn{1}{c}{NS}\\
\midrule
\multicolumn{22}{c}{$\Delta\Delta$LL $\sim$ SAS score}\\
\midrule
16M&\textcolor{darkgray}{0.21}&\textcolor{darkgray}{0.63}&\textcolor{darkgray}{-0.11}&\textcolor{darkgray}{0.46}&\textcolor{darkgray}{0.24}&\textbf{\textcolor{mdgreen}{0.7}}&\textcolor{darkgray}{0.09}&\textcolor{darkgray}{0.49}&.5B&\textcolor{darkgray}{-0.07}&\textcolor{darkgray}{0.15}&\textcolor{darkgray}{0.04}&\textcolor{darkgray}{0.09}&\textcolor{darkgray}{-0.14}&\textcolor{darkgray}{-0.09}&\textcolor{darkgray}{0.03}&\textcolor{darkgray}{0.1}&\textcolor{darkgray}{0.01}&\textcolor{darkgray}{-0.09}&\textcolor{darkgray}{-0.03}&\textcolor{darkgray}{-0.1}\\
32M&\textcolor{darkgray}{0.19}&\textcolor{darkgray}{-0.03}&\textcolor{darkgray}{-0.09}&\textcolor{darkgray}{0.47}&\textcolor{darkgray}{0.17}&\textcolor{darkgray}{-0.02}&\textbf{\textcolor{mdgreen}{0.72}}&\textcolor{darkgray}{0.02}&1B&\textcolor{darkgray}{-0.23}&\textcolor{darkgray}{-0.05}&\textcolor{darkgray}{-0.15}&\textcolor{darkgray}{-0.23}&\textbf{\textcolor{mdred}{-0.18}}&\textcolor{darkgray}{0.05}&\textcolor{darkgray}{-0.06}&\textbf{\textcolor{mdred}{-0.22}}&\textcolor{darkgray}{-0.06}&\textcolor{darkgray}{-0.04}&\textcolor{darkgray}{-0.01}&\textcolor{darkgray}{0.0}\\
64M&\textcolor{darkgray}{0.35}&\textcolor{darkgray}{0.56}&\textcolor{darkgray}{0.55}&\textcolor{darkgray}{0.45}&\textbf{\textcolor{mdgreen}{0.66}}&\textbf{\textcolor{mdgreen}{0.56}}&\textcolor{darkgray}{0.32}&\textcolor{darkgray}{0.25}&2B&\textcolor{darkgray}{-0.2}&\textcolor{darkgray}{0.17}&\textcolor{darkgray}{-0.13}&\textcolor{darkgray}{-0.23}&\textbf{\textcolor{mdred}{-0.25}}&\textcolor{darkgray}{-0.0}&\textcolor{darkgray}{0.0}&\textcolor{darkgray}{-0.15}&\textcolor{darkgray}{-0.04}&\textcolor{darkgray}{0.07}&\textcolor{darkgray}{-0.03}&\textcolor{darkgray}{-0.08}\\
128M&\textcolor{darkgray}{0.61}&\textcolor{darkgray}{0.39}&\textcolor{darkgray}{0.65}&\textcolor{darkgray}{0.5}&\textbf{\textcolor{mdgreen}{0.53}}&\textcolor{darkgray}{0.24}&\textcolor{darkgray}{0.33}&\textbf{\textcolor{mdgreen}{0.62}}&4B&\textcolor{darkgray}{-0.16}&\textcolor{darkgray}{-0.25}&\textcolor{darkgray}{0.16}&\textcolor{darkgray}{-0.23}&\textcolor{darkgray}{-0.13}&\textcolor{darkgray}{-0.0}&\textcolor{darkgray}{-0.02}&\textbf{\textcolor{mdred}{-0.23}}&\textcolor{darkgray}{0.02}&\textcolor{darkgray}{0.0}&\textcolor{darkgray}{-0.02}&\textcolor{darkgray}{-0.09}\\
256M&\textcolor{darkgray}{0.52}&\textcolor{darkgray}{0.62}&\textcolor{darkgray}{0.38}&\textcolor{darkgray}{0.49}&\textcolor{darkgray}{0.46}&\textcolor{darkgray}{-0.31}&\textcolor{darkgray}{0.1}&\textbf{\textcolor{mdgreen}{0.5}}&6B&\textcolor{darkgray}{-0.09}&\textcolor{darkgray}{-0.01}&\textcolor{darkgray}{0.17}&\textcolor{darkgray}{-0.02}&\textcolor{darkgray}{-0.05}&\textcolor{darkgray}{-0.02}&\textcolor{darkgray}{-0.0}&\textcolor{darkgray}{-0.03}&\textbf{\textcolor{mdgreen}{0.14}}&\textcolor{darkgray}{0.06}&\textcolor{darkgray}{-0.03}&\textcolor{darkgray}{0.06}\\
\midrule
\multicolumn{22}{c}{$\Delta\Delta$LL $\sim$ PS}\\
\midrule
16M&\textcolor{darkgray}{-0.1}&\textcolor{darkgray}{-0.69}&\textcolor{darkgray}{0.26}&\textcolor{darkgray}{-0.43}&\textcolor{darkgray}{0.41}&\textcolor{darkgray}{-0.36}&\textcolor{darkgray}{0.25}&\textcolor{darkgray}{0.08}&.5B&\textcolor{darkgray}{0.03}&\textcolor{darkgray}{-0.12}&\textcolor{darkgray}{-0.11}&\textcolor{darkgray}{-0.05}&\textcolor{darkgray}{0.07}&\textcolor{darkgray}{-0.03}&\textcolor{darkgray}{0.01}&\textcolor{darkgray}{0.01}&\textcolor{darkgray}{0.06}&\textcolor{darkgray}{0.02}&\textcolor{darkgray}{-0.01}&\textbf{\textcolor{mdgreen}{0.1}}\\
32M&\textcolor{darkgray}{-0.04}&\textcolor{darkgray}{0.14}&\textcolor{darkgray}{0.0}&\textcolor{darkgray}{-0.5}&\textcolor{darkgray}{-0.11}&\textcolor{darkgray}{0.06}&\textbf{\textcolor{mdred}{-0.51}}&\textcolor{darkgray}{-0.11}&1B&\textbf{\textcolor{mdgreen}{0.29}}&\textcolor{darkgray}{0.1}&\textcolor{darkgray}{0.13}&\textcolor{darkgray}{0.18}&\textcolor{darkgray}{0.15}&\textcolor{darkgray}{-0.0}&\textcolor{darkgray}{0.01}&\textcolor{darkgray}{0.06}&\textcolor{darkgray}{0.01}&\textcolor{darkgray}{0.05}&\textcolor{darkgray}{-0.02}&\textcolor{darkgray}{0.03}\\
64M&\textcolor{darkgray}{-0.33}&\textcolor{darkgray}{-0.61}&\textcolor{darkgray}{-0.56}&\textcolor{darkgray}{-0.43}&\textbf{\textcolor{mdred}{-0.64}}&\textcolor{darkgray}{0.31}&\textcolor{darkgray}{-0.05}&\textbf{\textcolor{mdred}{-0.77}}&2B&\textbf{\textcolor{mdgreen}{0.7}}&\textcolor{darkgray}{0.13}&\textcolor{darkgray}{0.16}&\textcolor{darkgray}{-0.11}&\textbf{\textcolor{mdgreen}{0.61}}&\textcolor{darkgray}{-0.0}&\textcolor{darkgray}{0.09}&\textcolor{darkgray}{0.15}&\textbf{\textcolor{mdgreen}{0.43}}&\textcolor{darkgray}{0.02}&\textcolor{darkgray}{0.03}&\textcolor{darkgray}{0.05}\\
128M&\textcolor{darkgray}{-0.53}&\textcolor{darkgray}{-0.25}&\textcolor{darkgray}{-0.53}&\textcolor{darkgray}{-0.6}&\textcolor{darkgray}{0.17}&\textcolor{darkgray}{0.48}&\textcolor{darkgray}{-0.23}&\textcolor{darkgray}{-0.5}&4B&\textbf{\textcolor{mdgreen}{0.6}}&\textcolor{darkgray}{0.13}&\textcolor{darkgray}{0.01}&\textcolor{darkgray}{0.26}&\textbf{\textcolor{mdgreen}{0.42}}&\textbf{\textcolor{mdred}{-0.23}}&\textcolor{darkgray}{0.13}&\textcolor{darkgray}{0.13}&\textbf{\textcolor{mdgreen}{0.21}}&\textcolor{darkgray}{0.03}&\textcolor{darkgray}{-0.06}&\textcolor{darkgray}{0.05}\\
256M&\textcolor{darkgray}{-0.46}&\textcolor{darkgray}{-0.53}&\textbf{\textcolor{mdred}{-0.74}}&\textcolor{darkgray}{-0.63}&\textcolor{darkgray}{0.21}&\textcolor{darkgray}{0.29}&\textcolor{darkgray}{-0.08}&\textcolor{darkgray}{-0.39}&6B&\textbf{\textcolor{mdgreen}{0.41}}&\textcolor{darkgray}{-0.09}&\textcolor{darkgray}{-0.2}&\textcolor{darkgray}{-0.06}&\textbf{\textcolor{mdgreen}{0.25}}&\textcolor{darkgray}{-0.13}&\textcolor{darkgray}{0.04}&\textcolor{darkgray}{0.08}&\textbf{\textcolor{mdgreen}{0.17}}&\textcolor{darkgray}{-0.02}&\textcolor{darkgray}{0.07}&\textbf{\textcolor{mdgreen}{0.12}}\\
\bottomrule
\end{tabular}

    \caption{Pearson correlation coefficients between SAS score and $\Delta\Delta$LL (top) and PS (bottom) and $\Delta\Delta$LL for 2 and 3 variants of GPT2 and Pythia, respectively. DU, ME, PR, and NS stand for Dundee, MECO, Provo, and Natural Stories, respectively. Correlations statistically significant at $\alpha=.05$ are boldfaced and colored in \textbf{\textcolor{mdgreen}{green}} if positive and \textbf{\textcolor{mdred}{red}} if negative. Note that these p-values are \textit{before} any correction for multiple testing is applied.}
    \label{tab:pearson}
\end{table*}
\subsection{Method}\label{sec:exp2:method}

\noindent\textbf{Scores.} PS is by definition a head-specific score (i.e. how much a given head attends to the previous occurrence of token B when given a sequence \aba). On the other hand, SAS is measured at the \textit{model} level; it is measured in UAS by picking attention scores of the best heads for each dependency relation type. Hence, we define a slightly modified \textit{head}-specific metric called \define{SAS score}. SAS score of the head $h$ at layer $l$ is a proportion of the words $w_i$ whose attention edge with the highest weight, $(i, \arg\max_{j} [a^{(h,l)}_{ij}])$, corresponds to a child-parent pair of a dependency relation.

\noindent\textbf{Ablation.} We zero-out attention weights of each head while keeping the original attention of all other heads and compute \dll (see \Cref{appendix:ablation} for the details). We then subtract the non-ablated score from the ablated score and call this value $\Delta\Delta$LL. A \textit{Higher} $\Delta\Delta$LL means that the ablation \textit{improves} PPP. We report the correlations between (1) SAS score and $\Delta\Delta$LL and (2) PS and $\Delta\Delta$LL.

\subsection{Results}\label{sec:exp2:result}
We start by considering a single, illustrative example: \Cref{fig:pms_dll} plots each head's PS ($x$-axis) and $\Delta\Delta$LL for Dundee corpus ($y$-axis) for Pythia-70M at 2B tokens of training. These particular results appear to confirm our hypothesis: heads high in PS are also high in $\Delta\Delta$LL, meaning that the heads performing induction \textit{damage} the PPP.

\begin{figure}[t]
    \input{latex/figures/pearsons.pgf}
    \vspace{-10pt}
    \caption{Relationship between each head's PS ($x$-axis) and the effect of its ablation on \dll ($y$-axis), measured on the Dundee orpus, for the Pythia-70M model at 2 billion pretraining tokens. Each point represents one of the 48 heads in the model.}
    \label{fig:pms_dll}
\end{figure}
However, we also need to test if this is the case for other models, checkpoints, and data (other than Dundee). We focus on 5 checkpoints around (and including) the phase transition point (64M for GPT2 models and 2B for Pythia models). This produces 5 (checkpoints) $\times$ 5 (models) $\times$ 4 (corpora) $\times$ 2 (scores: SAS score and PS) $=$ 200 scatterplots like \Cref{fig:pms_dll}. For readability, we only report Pearson's correlation coefficients for each, shown in \Cref{tab:pearson}. For example, \Cref{fig:pms_dll} can be found under column Pythia-70M and DU, row 2B on the bottom half of the table ($r=0.7$, $p<.05$).

In \Cref{tab:pearson}, a few trends emerge: First, we find strong correlations for models closer to the transition points (64M for GPT2 models and 2B for Pythia models). The very undertrained GPT2 models at 16M pretraining tokens, and Pythia models at 0.5B pretraining tokens, for example, show virtually no correlations between $\Delta\Delta$LL and SAS or PS. Second, whereas SAS$\sim$$\Delta\Delta$LL correlation is positive but PS$\sim$$\Delta\Delta$LL correlation is negative for GPT2 models, the opposite is true in general for Pythia models. We do not know why this should be the case, and it warrants further investigation. Given that more than half of the model checkpoints show no significant correlations, and that none of the significant correlations remain significant after Bonferroni correction for multiple testing is applied, we take these data as not supporting the claim that the post-phase transition degradation in PPP can be attributed to specific heads.

\section{Experiment 3: Suppressing Phase Transition Delays PPP Peak}\label{sec:exp3}

In this experiment, we test \Cref{hypo:2b} (Dynamics Hypothesis): if phase transition changes the course of pretraining dynamics, and if the post-transition pretraining dynamics lead to the degradation in PPP, suppressing the phase transition should result in the mitigation of PPP degradation.

\subsection{Methods}\label{sec:exp3:method}

Completely suppressing phase transition is difficult, if not impossible, as we do not have a full mechanistic understanding of this phenomenon. However, as induction heads and SAS are among the few well-documented diagnoses of phase transition, in this section we suppress them as a proxy method for suppressing phase transition more generally.

Since SAS quantifies attention heads whose attention patterns shadow dependency edges, \citet{chen2024sudden} propose using a syntactic regularizer:
\begin{equation}
\mathcal{L}_{\textsc{sas}}(x) = \underbrace{\mathcal{L}_{\textsc{clm}}(x)}_{\text{Original loss}} 
\underbrace{
    + \lambda \sum_{i=1}^{|x|} \sum_{x_j \in D(x_i)} \alpha(x_i, x_j)
}_{\text{Syntactic regularization}}
\end{equation}
where $x$ is an input, $D$ is a child-parent mapping of dependency relations, and $\alpha$ is an attention weight between a pair of words. $\lambda$ is a weighting factor, and \citet{chen2024sudden} find that $\lambda=0.001$ works best with BERT (positive $\lambda$ \textit{suppresses} SAS, whereas negative $\lambda$ \textit{promotes} it). We show results from $\lambda=\{0.01, 0.001\}$. We use \texttt{spaCy} \cite{honnibal2017spacy} to parse our training data.

To suppress the formation of induction heads, we regularize against attention patterns that correspond to the ``copying'' behavior:
\begin{equation}
\mathcal{L}_{\textsc{copy}}(x) = \underbrace{\mathcal{L}_{\textsc{clm}}(x)}_{\text{Original loss}} 
\underbrace{
    + \lambda\sum_{i=1}^{|x|} \sum_{x_j \in \text{PM}(x_i)} \alpha(x_i, x_j)
}_{\text{Copying regularization}}
\end{equation}
where $\text{PM}(x_i)$ is a prefix matching tokens of $x_i$. Recall that the copying behavior of induction heads was characterized by predicting \tokb when given a sequence \aba. Since we do not know which tokens are considered ``similar enough'' to promote the formation of induction heads in natural texts (see \citealp{chen-etal-2024-parallel} for the discussion on ``parallel structures'' that are central to LMs' ICL abilities), we construct synthetic data consisting of repeated random sequences of tokens (see \Cref{appendix:ih-reg} for the details of the dataset construction and copying regularization implementation).

In addition to the non-regularized model, copy-suppressed models, and SAS-suppressed models, we add models with Gaussian Noise Injection (GNI), as an example of a perturbed model with no explicit phase transition suppression. For more details, see \Cref{appendix:gni}.

\subsection{Results}\label{sec:exp3:result}

\subsubsection{Validation of Suppression}
We first verify that SAS and copying suppression are working as intended by showing that SAS suppression leads to a lower UAS and consequently a lower BLiMP score, and that copying impression leads to a near-zero PS and a reduced ICL score (these results are presented fully in \Cref{appendix:suppression}). 

Interestingly, however, none of the models resulted in a complete suppression of the phase transition. That is to say, with copy regularization, induction heads were suppressed (near-zero PS), but it promoted an even stronger SAS (higher UAS and BLiMP). On the other hand, with syntactic regularization, SAS was suppressed (near-zero UAS), but it delayed and weakened the induction heads (see \Cref{appendix:interaction} for more details on the interaction between the two types of regularization). All of the regularized models resulted in a higher loss compared to the non-regularized model, suggesting that emergent structures are indeed facilitative of next token prediction.
\begin{figure}[t]
    \hspace{-15pt}
    \input{latex/figures/splashy_with_suppression.pgf}
    \caption{Phase transition points ($x$-axis) and \dll tipping points ($y$-axis). Regularization delays both the phase transition and PPP tipping point, as exemplified by the SAS-regularized model ($\lambda$=0.01), shown with the red arrow. $\blacktriangle$ and \ding{72} represent phase transitions defined by UAS and PS, respectively. *$p <.05$; **$p < .01$.}

        \label{fig:splashy-with-suppression}
\end{figure}
\begin{figure*}[t]
    \input{latex/figures/pre_post_corr.pgf}
    \caption{Validation loss ($x$-axis) and \dll ($y$-axis) across checkpoints before (top) and after (bottom) the phase transition. Each point corresponds to a single checkpoint. The reported correlation is a Fisher $z$-transformed weighted average across models within each corpus. Correlation is consistently negative before the transition, while it becomes predominantly positive or mixed after. .$p<.1$; *$p <.05$; **$p < .01$; ***$p < .001$.}

        \label{fig:pre_post_corr}
\end{figure*}
\subsection{Effect of Supression on PPP}

Given that the phase transition was at most delayed (and not fully suppressed), and that loss is known to affect PPP, we present two sets of results: (1) the correlation between the phase transition points and PPP tipping points, and (2) the loss-PPP correlations before and after the phase transition. For (1), if the regularization resulted in a delayed phase transition, this should also result in a delayed PPP tipping point. Since PS is a better predictor of PPP tipping points (see \Cref{fig:splashy}), we use PS to compute phase transition points (\Cref{sec:exp1:method}). For models with copying regularization, UAS was used to obtain the phase transition points. For (2), we expect that the correlation is negative (i.e. quality-power hypothesis) before the phase transition, and positive after the phase transition. For the full trajectories of each model's PPP, see \Cref{appendix:details-suppression-ppp}.

\Cref{fig:splashy-with-suppression} summarizes the effect of regularization on phase transition points and PPP peaks, with the \textbf{\textcolor{Mahogany}{red arrow ($\rightarrow$)}} showing the representative case, pointing from the \textbf{\textcolor{RoyalPurple}{non-regularized model}} to \textbf{\textcolor{Dandelion}{SAS-suppressed model ($\lambda=0.01$)}}. MECO is again an outlier here; however, a strong correlation is found in both Dundee and NS, and also in Provo, to an extent. More importantly, as the arrow suggests, suppression that leads to a delayed phase transition also delays the PPP peak, moving the point along the \textbf{\textcolor{orange}{orange line}} that the Phase Transition Hypothesis predicts, and this further corroborates our \Cref{hypo:1}. For the magnitude and significance of these trends, see \Cref{tab:pre_post_corr} in \Cref{appendix:details-suppression-ppp}.

We have established that the suppression delays both phase transition and PPP tipping point. \Cref{fig:pre_post_corr} shows the loss-PPP correlation both before (top) and after (bottom) the phase transition. For Dundee and Provo, the contrast is evident: LM losses and PPPs have a negative correlation before the phase transition, and a positive correlation after it. The post-transition positive correlations we see especially in the Dundee and Provo corpora suggest that the decrease in loss that happens after the phase transition damages PPP, congruent with our Dynamics Hypothesis (\Cref{hypo:2b}).
\section{Discussion}\label{sec:discussion}
This work proposed phase transition as the underlying factor that causes PPP degradation beyond a certain point during LM pretraining and conducted several studies to test this hypothesis. Here, we discuss how our results relate to previous factors that have been shown to impact PPP in LMs.
\subsection{Phase Transition \& Factors Affecting PPP}\label{sec:discussion:factors}
First, our Phase Transition Hypothesis can potentially explain previous findings that limiting the context window of a transformer LM can improve PPP \cite{kuribayashi-etal-2022-context}, especially for infrequent tokens \cite{oh-etal-2024-frequency}, and that adding a linear recency bias improves PPP \cite{de-varda-marelli-2024-locally, clark-etal-2024-linear}. To explain why, recall that \citet{olsson2022context} find that induction heads improve the ICL score, as measured by how well an LM leverages earlier tokens in the context for the next word prediction, even across long spans of text. If the emergence of ICL abilities is the underlying cause of the PPP reversal, then it follows that limiting ICL would help PPP. In this study, we have achieved this through ablation and regularization; however, restricting LM's context size would yield a similar outcome by completely disabling models to leverage earlier tokens.

\subsection{Phase Transition \& ``Bigger $\ne$ Better''}\label{sec:discussion:biggerbetter}

Second, the Phase Transition Hypothesis can explain why larger models suffer from lower PPP than smaller models (e.g., \citealp{oh-etal-2022-gpt, oh-schuler-2023-surprisal}), an example of inverse scaling.  \citet{oh-schuler-2023-surprisal} conjecture that the overly accurate prediction of infrequent tokens is one of the contributing factors to models' PPP reversal, and suggest that it could be due to memorization. \citet{chen2024sudden} suggest that the memorization phase occurs after the phase transition. Taken together, given that most of the models studied in previous works have almost certainly undergone the phase transition, it is unsurprising that memorization is damaging PPP. In fact, \citet{oh-etal-2024-frequency} make an interesting observation regarding pretraining dynamics and predictions on infrequent tokens: models of different sizes seem to follow similar learning trajectories up to a certain point during pretraining (2B tokens), but the ability to predict infrequent tokens with low surprisals (which is considered one of the reasons for the larger models' poor predictive power) seems to emerge beyond that point only among larger models.

\subsection{SAS \& Induction Heads}\label{sec:discussion:sas-ih}

To our knowledge, this study is the first to investigate SAS in decoder-only models, and to observe the emergence of induction heads and SAS in the same model. This affords us to make observations about the interactions between the two, as briefly discussed in \Cref{sec:exp3:result}.

In all of the models we examined, SAS emerges first, closely followed by the formation of induction heads (see \Cref{fig:coinciding} in \Cref{appendix:full-results}). Although the universality of this order remains inconclusive based on the selection of models we examined, this begs a question: is the emergence of SAS a prerequisite for the emergence of induction heads? 

If this was the case, we would expect the SAS suppression to also suppress induction head formation, but not vice versa. Interestingly, we see a similar trend, but not quite what we expected: suppression of SAS delays the induction head formation and lowers the eventual PS obtained at the end of the training; however, the suppression of induction heads does not affect the point at which SAS emerges, and further, it \textit{increases} the UAS. The interaction among different types of specialized heads remains an interesting topic for further investigation (see \Cref{appendix:interaction} for details).

\subsection{Beyond Reading Time}\label{sec:discussion:beyond}

Concurrent works explore the relationship between training dynamics and the alignment between LMs and brain activities. \citet{alkhamissi2025languagecognitionllmsoutgrow} measured Pythia models' formal and functional linguistic competences as well as their alignment with brain data throughout pretraining and reported a strong correlation between formal linguistic competence and the brain-LM alignment. Furthermore, they show that models of various sizes (ranging from 70M--6.9B) have similar levels of alignment with brain activities, and that they all tend to peak in the alignment between 2B--8B tokens of pretraining, echoing the observation we have made. Note that the tipping point of brain-LM alignment seems to come after the tipping point of sentence processing alignment.

\citet{nakagi2025triple} find a similar abrupt increase in the brain-LM alignment at 1B pretraining tokens, followed by a sudden drop at 10B pretraining tokens for a set of OLMo variants. They find a subsequent resurgence in the alignment score at around 50B pretraining tokens, describing the overall dynamics as consisting of 3 phases. Although we only trained our models up to 10B tokens, results from \citet{oh-schuler-2023-transformer} show that Pythia models go through a tipping point at around 2B tokens, after which no further increase is observed throughout the 300B pretraining tokens. Hence, this resurgence after the tipping point might be unique to brain-LM alignment; however, more work is needed to verify this speculation.

\section{Conclusion}

This study found a strong correlation between the phase transitions and peaks in a model's fit to human reading times. However, the precise aspect of the phase transition responsible for PPP degradation remains somewhat inconclusive. Given that the results largely supported \Cref{hypo:2b} (Dynamics Hypothesis, \Cref{sec:exp3}) rather than \Cref{hypo:2a} (Persistence Hypothesis; \Cref{sec:exp2}), we can be reasonably confident that it is the post-phase transition dynamics driving the PPP degradation. However, it is important to note that none of our regularization methods was able to completely suppress the phase transition. Fully controlling the phase transition and investigating its effect on PPP requires a more complete understanding of the phase transition, and this remains an important avenue for future research.

\section{Limitations}\label{sec:limitation}

First, our models are limited to Pythia models and GPT2 models. This is due to the limited availability of pretrained models' checkpoints, as well as the high computational cost associated with training large models from scratch. As mentioned in \Cref{sec:method:models}, to our knowledge, Pythia is the only model family whose available checkpoints cover the phase transition points.

We also limited the GPT2 models to 0-, 1-, and 2-layer variants. This is because these three variants are theoretically shown to behave qualitatively differently \cite{olsson2022context} as we discussed in \Cref{sec:exp1:result}, and that models with 2 or more layers are not qualitatively different for our purpose of attention-based phase transition detection. Pythia model's 70M, 160M, and 410M variants are 6, 12, and 24 layers, respectively, and we believe that this covers a reasonable range of model sizes, together with the GPT2 models we train. We also note that experiments on smaller ``toy'' models are important first steps (e.g., \citealp{elhage2021mathematical, olsson2022context, power2022grok, kallini-etal-2024-mission}). Yet, training larger models in \Cref{sec:exp3} was beyond our compute budget, and we leave this to future work.

Second, our selection of reading time corpora is representative rather than comprehensive, and the replication with other corpora, as well as different reading behavior metrics such as brain activity data, remains an important future work. We also observe that the results for MECO were different from results from the other 3 corpora. The absence of the PPP peak $\sim$ phase transition correlation (\Cref{fig:splashy} in \Cref{sec:intro} and \Cref{fig:splashy-with-suppression} in \Cref{sec:exp3:result}) are considered an exception rather than a rule. Several issues that could affect the quality of the corpus have been reported on MECO. For example, \citet{opedal-etal-2024-role} find an off-by-one issue for a handful of tokens in MECO, as well as repeated words in a few sentences. We followed the fixed version of the data; however, there may be other issues we are unaware of, potentially causing the divergent behavior of the corpus.

Lastly, our study is limited to English, and our results may not hold for other languages. However, given that syntactic dependencies and word or phrase repetitions are universal across the world's languages, we predict that a similar trend might be observed in other languages. A multilingual extension of this study is therefore a promising direction for future work.

\section*{Ethics Statement}
We trained several small transformer-based LMs from scratch, which could contribute to the increased carbon footprint. However, we train models that have at most 2 layers and 8 heads. By choosing a model size that reasonably approximates the popular transformer architecture (at least for the purpose of our study) while curtailing the computational cost, we believe that we were able to minimize our environmental impact.

Lastly, while our study utilizes four sources of human behavioral data, we do not intend to redistribute or publicly share these datasets. We affirm that our use of this data aligns with ethical standards and does not pose any potential ethical concerns.

\section*{Acknowledgments}

We thank the four anonymous reviewers for their active engagement in the discussion, which made the final version of this paper substantially better.

\bibliography{final}

\appendix

\section{Hyperparameters}\label{sec:appendix-hyperparams}

\begin{table}[h]
    \centering
    \begin{tabular}{cr}
\toprule
\multicolumn{2}{c}{Architecture}\\
\midrule
\texttt{architecture} & \texttt{gpt2} \\ 
\texttt{vocab\_size} & \texttt{50\_257} \\ 
\texttt{context\_size} & \texttt{1\_024} \\
\texttt{d\_embed} & \texttt{768}\\
\texttt{d\_ffn} & \texttt{3\_072}\\
\texttt{n\_layer} & \texttt{2}\\
\texttt{n\_head} & \texttt{8}\\
\texttt{activation} & \texttt{gelu} \\ 
\texttt{num\_params} & \texttt{53M} \\ 
\midrule
\multicolumn{2}{c}{Training}\\
\midrule
\texttt{train\_size} & \texttt{1B} \\
\texttt{num\_epoch} & \texttt{10}\\ \texttt{train\_amount} & 
\texttt{10B} \\
\texttt{batch\_size} & \texttt{4}\\
\texttt{grad\_acc\_steps} & \texttt{1}\\
\texttt{weight\_decay} & \texttt{0.1}\\
\texttt{warmup\_steps} & \texttt{1\%}\\
\texttt{lr} & \texttt{5e-4}\\
\texttt{lr\_scheduler} & \texttt{cosine}\\

\bottomrule
\end{tabular}
    \caption{List of hyperparameters used to train L2LMs.}
    \label{tab-hyperparams}
\end{table}

\Cref{tab-hyperparams} summarizes the hyperparameters used to train 2-layer GPT2 models. 0-layer and 1-layer GPT2 models had 39M and 46M total parameters, respectively, and the rest of the hyperparameters remain the same.

\section{Regression Model Formulae for \dll Calculation}\label{appendix:ppp}
Following \citet{SMITH2013302}, we model the spillover effect of previous 2 and 4 words for modeling eye-tracking data (Dundee, MECO, Provo) and self-paced reading data (Natural Stories), respectively. Frequencies are estimated using Wikitext.

\noindent\textbf{Baseline regression model for eye-tracking.}
\texttt{psychometric $\sim$ freq + prev\_freq + prev2\_freq + len + prev\_len + prev2\_len}

\noindent\textbf{Baseline regression model for reading time.}

\noindent\texttt{psychometric $\sim$ freq + prev\_freq + prev2\_freq + prev3\_freq + prev4\_freq + len + prev\_len + prev2\_len + prev3\_len + prev4\_len}

\noindent\textbf{Full regression model for eye-tracking.}

\noindent\texttt{psychometric $\sim$ surprisal + prev\_surp + prev2\_surp + freq + prev\_freq + prev2\_freq + len + prev\_len + prev2\_len}

\noindent\textbf{Full regression model for reading time.}

\noindent\texttt{psychometric $\sim$ surprisal + prev\_surp + prev2\_surp + prev3\_surp + prev4\_surp + freq + prev\_freq + prev2\_freq + prev3\_freq + prev4\_freq + len + prev\_len + prev2\_len + prev3\_len + prev4\_len}

\section{Unlabeled Attachment Score (UAS)}\label{appendix:uas}

We follow the calculation of UAS introduced in \citet{chen2024sudden}, which is based on \citet{clark-etal-2019-bert}. While they both use bidirectional models (i.e. BERT), we use decoder-only autoregressive models (i.e. GPT2 and Pythia), and hence make modifications to account for this difference. The overall recipe is to (1) define a head-specific probe, (2) find the best performing head for each dependency relation type, and (3) calculate the overall UAS using the best heads defined in (2).

\noindent\textbf{(1) Head-specific probe.} A head-specific probe $f_{h,l}$ predicts the parent word of a target word $x_i$ by selecting the word $x_j$, whose attention edge to or from the target word $x_i$ is the highest among all words $x_j \in \{j\ne i\}$ for a given head $h$ at layer $l$:
\begin{equation}
    f_{h,l}(x_i) = \arg \max_{x_j} \left( a^{(h,l)}_{ij}\right).
\end{equation}
Note that, unlike bidirectional models, where each pair of words is connected by 2 attention edges ($x_i\leftrightarrow x_j)$, only 1 edge lies between any pair of words for autoregressive models. This means that the number of words a given word $x_i$ \textit{can} attend to is $i$, whereas the average number of words $x_i$'s right context \textit{can} attend to is $\frac{i+1024}{2}$, creating discrepancy in the scale. If $i=10$, for example, with the context size of 1024, $\forall_{j\le i}[\mathbb{E}(a^{(h,l)}_{i\rightarrow j})]=0.1$, whereas $\mathbb{E}_{j>i}[\mathbb{E}(a^{(h,l)}_{j\rightarrow i})]\approx0.002$. However, we find that scaling the attention weights between $a^{(h,l)}_{i\rightarrow j}$ and $a^{(h,l)}_{j\rightarrow i}$ produce similar results, and hence we report unscaled results throughout.

Following \citet{clark-etal-2019-bert, chen2024sudden}, we convert the token-level attention to word-level attention by summing over attention weights \textit{to} destination tokens that make up a single word, and by averaging over attention weights \textit{from} source tokens that constitute a single word.

\noindent\textbf{(2) Best head per relation type.} The rest of the UAS calculation remains same as \citet{chen2024sudden}. We now convert the head-specific probe defined in (1) to a relation-specific probe by finding the best head for each dependency relation type.

For each dependency relation type $R$, which we define as a set of all ordered child-parent pairs $(x, y)$, the best performing head for the given dependency relation type is:
\begin{equation}
    \hat{f}_R = \arg \max_{f_{h,l}} \frac{1}{|R|} \sum_{(x, y) \in R} \mathbbm{1}_R \left( x, f_{h,l}(x) \right),
\end{equation}
where $x$ and $y$ are constrained to be within the same sentence. The indicator function for set $R$, $\mathbbm{1}_R$ is 1 if the predicted child-parent pair is in the set $R$, and 0 otherwise. Hence, $\hat{f}_R$ is simply a head that has the highest recall for a given dependency relation type $R$.

\noindent\textbf{(3) UAS.} Lastly, we simply take the average of the performance of each relation type's best head over all relation types, weighted by the number of ordered word pairs in that relation type. Denoting the set of all relation types as $\mathcal{R}$, UAS is defined as:
\begin{equation}
    \text{UAS} = \frac{1}{\sum_{R \in \mathcal{R}} |R|} \sum_{R \in \mathcal{R}} \sum_{(x_i, x_j) \in R} \mathbbm{1}_R \left( x_i, \hat{f}_R(x_i) \right).
\end{equation}

\section{In-Context Learning (ICL) Score}

\citet{olsson2022context} define what they call \textbf{ICL score}, or the difference in the losses of tokens later in the context and tokens earlier in the context. This score seems to be robust to the choice of exactly which tokens to compare; however, \citet{olsson2022context} report the difference between 50th and 500th tokens' losses. We instead report the difference between the average loss of early tokens $w_j$, where $j\in[40,60]$, and that of later tokens $w_k$, where $k\in[450,550]$, for each sequence $\mathbf{s}_i$, where $i\in[1, N]$:
\begin{equation}\label{eq:icl-score}
\frac{
\sum_{i}
    \sum_{j} \sum_{k} \mathcal{L}(f_\theta(\mathbf{s}_i, w_{j})) - \mathcal{L}(f_\theta(\mathbf{s}_i, w_{k}))}{200\times N}
\end{equation}
where $f_\theta(s_i, w_{j})$ is the output of an LM parametrized by $\theta$,  given $j$-th word $w_j$ in $i$-th sequence $\mathbf{s}_i$.
A positive ICL score means that the model has a lower loss (i.e. better prediction) later in the context than earlier in the context.

\section{Phase Transition and its Downstream Effects}\label{appendix:pt-details}

\begin{figure}[t]
    \hspace{-15pt}
    \input{latex/figures/UAS_BLiMP.pgf}
    \vspace{-20pt}
    \caption{Trajectories of BLiMP score over 10B tokens of pretraining. $\blacktriangle$ represent phase transition defined by a sudden rise UAS.}
    \label{fig:uas-blimp}
\end{figure}

\citet{chen2024sudden} find that the increase in UAS triggers the acquisition of syntactic abilities, as shown in an increase in the BLiMP score closely following the UAS boost, and we find a similar pattern. In \Cref{fig:uas-blimp}, GPT2 seems to go through a drastic increase in the BLiMP score between 4M to 1B pretraining tokens, with a brief halt around 32M-64M pretraining tokens. This is when the phase transition occurs, and it seems to signal the onset of the second boost in BLiMP score, starting around 64M tokens. For Pythia, the picture seems even clearer: UAS phase transition happens at around 512M pretraining tokens, immediately followed by a sudden increase in the BLiMP score.

Similarly, \citet{olsson2022context} reports a dramatic improvement in ICL score, foreshadowed by the emergence of induction heads, measured by PS. In \Cref{fig:pms-icl}, ICL scores of 2-layer GPT2 and 1-layer GPT2 models seem to start diverging between 32M-64M pretraining tokens, which is exactly when the 2-layer model is undergoing the phase transition. Recall that induction requires attention composition, which is only possible with models with 2 or more layers. For Pythia models, the increase in ICL score is most dramatic between 1B and 2B pretraining tokens, which also coincide with the emergence of induction heads.

\section{Ablation}\label{appendix:ablation}

Ablation of attention heads can be implemented in two ways: full ablation and pattern-preserving ablation. For full ablation, the attention output is simply set to 0. Recall that attention output is a matrix multiplication between the attention weight vector, which is computed using query and key vectors, and the value vector \cite{vaswani-etal-2017-attention}:
\begin{equation}\label{eq:attention}
    \mathrm{Attn}(Q, K, V)=\mathrm{softmax}(\frac{QK^T}{\sqrt{d_k}})V,
\end{equation}
where Q, K, and V are query, key, and value vectors, respectively, and $d_k$ represents the dimension of Q and K vectors.
Because Q, K, and V vectors are linear projections of the previous layer's output, setting \Cref{eq:attention} to a 0 vector affects all downstream (later layers') calculations of Q, K, and V vectors.
\begin{figure}[t]
    \hspace{-25pt}
    \input{latex/figures/PS_ICL.pgf}
    \vspace{-20pt}
    \caption{Trajectories of ICL score over 10B tokens of pretraining. \ding{72} represent phase transition defined by a sudden rise PS.}
    \label{fig:pms-icl}
\end{figure}
\begin{figure*}[t]
    \hspace{-30pt}
    \input{latex/figures/coinciding.pgf}
    \vspace{-20pt}
    \caption{Trajectories of \dll over 10B tokens of pretraining. $\blacktriangle$ and \ding{72} represent phase transition defined by a sudden rise in UAS and PS, respectively.}
    \label{fig:coinciding}
\end{figure*}
In pattern-preserving ablation, we feed the input to the model twice: during the first run, no heads are ablated, and we simply record all the attention weights at each head of each layer. During the second run, we ablate the head(s) of interest, but use the attention weights recorded in the first run, preserving the original attention weights (hence \textit{pattern-preserving}). Because only $Q$ and $K$ vectors are involved in the attention weight calculation $\mathrm{softmax}(\frac{QK^T}{\sqrt{d_k}})$, pattern-preserving ablation only affects downstream $V$ calculations, but not $QK$ calculations. We note that these two implementations of ablation yield similar results, and we report the results obtained from pattern-preserving ablation.

\section{Full PPP Trajectory and Phase Transition Points}\label{appendix:full-results}

\Cref{fig:coinciding} shows a fuller visualization of the entire trajectory of \dll, with the phase transition points marked for each model ($\blacktriangle$ and \ding{72} for UAS and PS, respectively, which correspond to the x-axis of \Cref{fig:splashy}). Note that, in \Cref{fig:coinciding}, we included 0 and 1 layer GPT2 models as they are expected to behave differently from 2 layer models. This is because 0 layer model does not have attention layers, and 1 layer model cannot form induction heads, as they require \textit{attention composition} \cite{olsson2022context}. We limited the 2 layer models to the one trained with a batch size of 4 for readability. 

First, GPT2 with no attention layer (\textcolor{violet}{dark purple line}) exhibits no ``tipping point'' in PPP, meaning that it never seems to undergo a point at which PPP starts going down. This is expected as the definitions of phase transition we adopt in this paper are both triggered by specialized behaviors of attention heads, and this further corroborates our hypothesis. Conversely, for GPT2 models with 1 and 2 layers, as well as 3 different sizes of Pythia models all have a tipping point at which PPP starts going down. Note that the results for Pythia models are partial replications of \citet{oh-schuler-2023-transformer}.

Second, as shown in \Cref{fig:splashy} as well, the tipping points of PPP seem to closely match the breakthrough points in ICL scores. Note that these breakthrough points are meaningful for models with 2 or more layers because induction heads work with another head in lower layers to perform the copying behavior (cf. attention composition; \citealp{olsson2022context}); hence, we do not indicate those breakthrough points for 0 and 1 layer models.
\section{Implementation of Copying Regularization}\label{appendix:ih-reg}

Since the distance between two \ab sequences, $l$, can be arbitrarily large within the LM's context size, we sample $l$ from a uniform distribution $U(50, 512)$. The lower bound is to make sure non-copying behaviors do not get suppressed, and the upper bound is just half the max context length of our model. To maximize the suppression target, we repeat a random sequence of length $l$ as many times as it takes to fill the context size of 1024. Since it has to be repeated at least twice to promote the copying behavior, $l$ is upper-bounded by 512. With this synthetic data, $PM(x_i)$ can be written as:
\begin{equation}
\{x_j|j<i, x_{j}=x_{i+1}, \mathbf{x}_{i-l+1}^i=\mathbf{x}_{j-l}^{j-1}\}
\end{equation}
or programmatically:
\begin{equation}
\{x_{i-nl+1}|n\in\mathbb{N}_1, x-nl>0\}
\end{equation}
Now, as opposed to SAS suppression, where the regularization term is computed on the same set of examples from which normal loss $\mathcal{L}_{\textsc{clm}}$ is computed, the copying regularization term requires separate synthetic data. We considered alternating between the $\mathcal{L}_{\textsc{clm}}$ and $\mathcal{L}_{\textsc{copy}}$ every $\frac{1}{\lambda}$ steps; however, we find that this leads to a very unstable learning curve, and hence add the regularization term every step weighted by $\lambda$ as was the case with syntactic regularization.
\section{Permutation Test}\label{appendix:permutation}
To test the significance of the differences in \dll between non-regularized and regularized models, we conduct a series of pair-wise permutation tests. Since \dll is computed for each word per model, and we are not interested in the variance in \dll introduced by word ID, it was treated as a random effect. More concretely, for each pair of models we are comparing, we first generate 10,000 random permutations (i.e. random assignment of group) between the 2 labels (2 models being compared) for \textit{each word ID}, so that each of the per-word \dll is randomly assigned to one of the two labels. We then count the proportion of permutations that have a mean within-word ID difference equal to or larger than the observed difference. Since we are interested in the difference between non-regularized and regularized models, we conduct six pair-wise permutation tests between the non-regularized model and each of the six regularized variants ($3\text{ regularizers}\times2\text{ hyperparameters}=6$). \Cref{tab:pvals} summarizes the permutation test results. Note that the p-values are all \textit{before} any correction for multiple testing.
\begin{table}[t]
    \centering
    \small\begin{tabular}{lr|rrrr}
\toprule
\multicolumn{2}{c|}{\textbf{Model}}&\multicolumn{4}{c}{\textbf{Corpora}}\\

Reg & $\lambda$/$\sigma$ &Dundee & MECO & Provo & NS\\
\midrule
\multirow{2}{*}{Copy} & 0.001 & .334 & .539 & .000 & .986 \\

 & 0.01 & .076 & .269 & .003 & .271 \\

\multirow{2}{*}{Syntax} & 0.001 & .305 & .808  & .001 & .194 \\

 & 0.01 & .121 & .947 & .319 & .023 \\
 
\multirow{2}{*}{GNI} & 0.05 & .461 & .175 & .264 & .539 \\

& 50 & .173 & .299 & .301 & .016 \\

\bottomrule
\end{tabular}
    \caption{P-values obtained from the permutation test.}
    \label{tab:pvals}
\end{table}
\section{Regularization}\label{appendix:suppression}
\subsection{UAS and BLiMP}
\begin{figure}[t]
    \hspace{-10pt}
    \input{latex/figures/UAS_reg.pgf}
    \caption{Trajectories of UAS over 10B tokens of pretraining with and without SAS suppression. All models are GPT2 with 2 layers with 8 attention heads.}
    \label{fig:uas-reg}
\end{figure}
\begin{figure}[t]
    \hspace{-10pt}
    \input{latex/figures/BLiMP_reg.pgf}
    \caption{Trajectories of BLiMP score over 10B tokens of pretraining with and without SAS suppression. All models are GPT2 with 2 layers with 8 attention heads.}
    \label{fig:blimp-reg}
\end{figure}

\Cref{fig:uas-reg} summarizes the development of UAS for the GPT2 models with and without SAS suppression. As opposed to the non-suppressed model (NoReg), whose UAS abruptly increases between 16M and 64M pretraining tokens, the SAS-suppressed model with $\lambda=0.001$ exhibits a brief increase in UAS, followed by a gradual degradation, converging to almost 0 towards the end of the pretraining. The SAS-suppressed model with stronger suppression, $\lambda=0.01$, almost never sees any improvement in UAS throughout the pretraining. We confirm that the SAS suppression is working as intended.

\Cref{fig:blimp-reg} summarizes the development of BLiMP score over the course of pretraining for the same 3 models discussed above. After 16M to 64M pretraining tokens, when the phase transition was \emph{supposed to} happen, BLiMP score perfectly correlates with the SAS suppression strength, with non-regularized model performing the best, and the strongly regularized model performing the worst.

\begin{figure}[t]
    \hspace{-10pt}
    \input{latex/figures/PS_reg.pgf}
    \caption{Trajectories of PS over 10B tokens of pretraining with and without copying suppression. All models are GPT2 with 2 layers with 8 attention heads.}
    \label{fig:pms-reg}
\end{figure}

\subsection{PS and ICL}

\Cref{fig:pms-reg} plots the development of best PS. Recall that PS is a head-level score, and we show the score from the highest scoring head for each checkpoint. At 64M pretraining tokens, we see the phase transition in non-regularized model. Both copying-suppressed models ($\lambda=0.001, \lambda=0.01$) show almost no improvement in the best PS throughout the course of pretraining.

\begin{figure}[t]
    \hspace{-10pt}
    \input{latex/figures/ICL_reg.pgf}
    \caption{Trajectories of ICL score over 10B tokens of pretraining with and without copying suppression. All models are GPT2 with 2 layers with 8 attention heads.}
    \label{fig:icl-reg}
\end{figure}
\begin{figure}[t]
    \hspace{-20pt}
    \input{latex/figures/interaction_ps.pgf}
    \caption{Trajectories of PS over 10B tokens of pretraining. All models are GPT2 with 2 layers with 8 attention heads.}
    \label{fig:interaction-ps}
\end{figure}
\begin{figure}[t]
    \hspace{-20pt}
    \input{latex/figures/interaction_icl.pgf}
    \caption{Trajectories of ICL over 10B tokens of pretraining. All models are GPT2 with 2 layers with 8 attention heads.}
    \label{fig:interaction-icl}
\end{figure}
\begin{figure}[t]
    \hspace{-10pt}
    \input{latex/figures/interaction_uas.pgf}
    \caption{Trajectories of UAS over 10B tokens of pretraining. All models are GPT2 with 2 layers with 8 attention heads.}
    \label{fig:interaction-uas}
\end{figure}
\begin{figure}[t]
    \hspace{-10pt}
    \input{latex/figures/interaction_blimp.pgf}
    \caption{Trajectories of BLiMP over 10B tokens of pretraining. All models are GPT2 with 2 layers with 8 attention heads.}
    \label{fig:interaction-blimp}
\end{figure}
\Cref{fig:icl-reg} shows that the copying-suppressed models indeed show lower ICL scores, presumably as a result of the absence of induction heads. Notably, at around 64M pretraining tokens, when phase transition occurs in the non-regularized model, both regularized models see a large dip in the ICL score. This is perhaps because the emergent copying behavior briefly raises the penalizing term of the loss function $\mathcal{L}_{\textsc{copy}}$, resulting in a brief phase of negative ICL scores.

\section{Interaction}\label{appendix:interaction}
We find that suppressing one type of specialized head affects the behavior of the other type of specialized heads. As briefly discussed in \Cref{sec:discussion:sas-ih}, because SAS always emerges \textit{before} induction heads, it is expected that the suppression of SAS affects the performance of induction heads, but not vice versa.

\subsection{SAS supression $\rightarrow$ induction heads} In \Cref{fig:interaction-ps}, we can see that the suppression of SAS indeed leads to slower emergence and weaker performance of induction heads, and this trend is more pronounced for a stronger regularizaiton, where $\lambda=0.01$. Consequently, in \Cref{fig:interaction-icl}, SAS suppression makes the improvement in ICL slower, and the eventual ICL lower.
\begin{figure}[t]
    \hspace{-10pt}
    \input{latex/figures/reg_effect_10b_with_GNI.pgf}
    \caption{\dll of 7 variants of 2 layer GPT2 models at the end of pretraining (10B tokens). Red dotted lines represent the \dlls obtained from the non-regularized GPT2. Within each corpus, regularized models are compared against the non-regularized model (dotted lines) to test statistical significance. Note that p-values are before applying any correction for multiple testing.\\$p>.1$ (ns); $p < 0.1$ (.); $p < 0.05$ (*); $p < 0.01$ (**)}
    \label{fig:10b_with_GNI}
\end{figure}
\begin{figure*}[t]    
    \hspace{-20pt}
\input{latex/figures/reg_trajectory.pgf}
    \caption{Effect of regularization on \dll. All models are 2-layer GPT2 models, trained with a batch size of 4.}
    \label{fig:suppression_dll_all}
\end{figure*}

\subsection{Induction head supression $\rightarrow$ SAS} On the other hand, as expected, \Cref{fig:interaction-uas} show that the suppression of induction heads does not suppress the emergence of SAS; rather, surprisingly, it \textit{improves} UAS for both values of $\lambda$. Consequently, with the stronger attention to dependency edges, models with induction head suppression performs better in BLiMP (\Cref{fig:interaction-blimp}).
\subsection{Asymmetry between SAS and Induction Heads}
Why does the suppression of SAS \textit{delays and weakens} the formation of induction heads, whereas the suppression of induction heads \textit{strengthens} SAS? Given that SAS always emerges before induction heads, one possible explanation is that the presence of natural grammar (the fact that paying strong attention to a token connected with a dependency edge facilitates next token prediction) may have a scaffolding effect for learning non-uniform and highly concentrated attention distribution, with the majority of the mass allocated to a single token. Hence, the removal of this scaffolding (i.e. SAS suppression) delays the emergence of induction heads. Note that the strong SAS suppression, with almost 0 UAS throughout (yellow line in \Cref{fig:interaction-uas}), still produces some level of PS (yellow line in \Cref{fig:interaction-ps}). This implies that SAS is a mere scaffolding for induction head formation, and not a prerequisite.

On the other hand, since SAS emerges before induction heads, SAS seems not to be using such a scaffolding from the formation of induction heads. Rather, in natural language with \abab sequences, induction heads may be \textit{inhibiting} the model's ability to attend to tokens connected with dependency edges, by allocating some of the attention weights, that \textit{could have been} used for SAS, to prefix-matching tokens. Note that, in our copy suppression, prefix-matching tokens were at least 50 tokens apart from each other, making dependency-connected tokens and prefix-matching tokens disjoint sets for most cases. However, these are pure speculations, and further studies are needed to confirm exactly why the specialized heads' training dynamics interact in the way we observed.

\section{Gaussian Noise Injection}\label{appendix:gni}

The models with Gaussian Noise Injection (GNI) were trained on the same dataset as other models (CC100), while adding to FFNs noises sampled from a Gaussian distribution with the mean of 0:
\begin{equation}
    \tilde{h} = h + \epsilon, \quad \epsilon \sim \mathcal{N}(0, \sigma^2)
\end{equation}
where $h$ is the activation of a transformer block's FFN. As $\sigma$ determines the magnitude of the noise, we included models with $\sigma=0.05$ and $\sigma=50$.

\section{Details of the Effect of Suppression on PPP}\label{appendix:details-suppression-ppp}

In \Cref{sec:exp3:result}, we mainly reported (1) the correlation between phase transition points and PPP tipping points, and (2) loss-PPP correlations before and after the phase transition. This is because PPP is known to be a function of loss (\textit{quality-power hypothesis}; \citealp{goodkind-bicknell-2018-predictive}) as well as whether the model has gone through the tipping point (or equivalently, the phase transition, based on our hypothesis). Hence, comparing the absolute PPPs is not necessarily informative: model A having a higher PPP than model B can mean (1) both are before the phase transition and A has a lower loss, or (2) both are after the phase transition and A has a higher loss, among other possibilities. Nonetheless, for the sake of completeness, we present the absolute differences, as well as their statistical significance (\Cref{fig:10b_with_GNI}), and the full trajectories of PPP (\Cref{fig:suppression_dll_all}).

In \Cref{fig:10b_with_GNI}, first, we find that syntactic regularization tends to improve PPP over non-regularized models, and this is robust to the settings of $\lambda$ or the corpus. Second, syntactic regularization is also more effective at improving PPP than copying regularization, which shows mixed results; its impact is negative at the lower lambda value ($\lambda=0.001$) for the Natural Stories corpus, whereas for MECO, it adversely affects PPP regardless of the regularization strength. See \Cref{appendix:permutation} for details on the permutation tests for statistical significance.

In \Cref{fig:suppression_dll_all}, the degradation in PPP at the phase transition point is partly suppressed, although the results vary by regularization type, strength, and corpus. For example, for the Dundee corpus, syntactically regularized models have lower PPP around the transition point compared to the original model; however, they degrade less than the original model and consequently end up with higher PPP later in pretraining. This trend is much more pronounced in the Natural Stories Corpus; as opposed to the original model, whose PPP decreases after the transition point, SAS-suppressed models either plateau ($\lambda=0.001$) or keep improving ($\lambda=0.01$) beyond that point. However, the degradation trend is not fully suppressed or reversed for many models and corpora.
\begin{figure}[t]
    \hspace{-10pt}
    \input{latex/figures/loss.pgf}
    \caption{Trajectories of validation loss on over 10B tokens of pretraining. All models are GPT2 with 2 layers with 8 attention heads.}
    \label{fig:loss}
\end{figure}
\begin{table*}[t]
    \centering
    \small
    \begin{tabular}{ll|rrrrrrr}
\toprule
&\textbf{Corpus}&\multicolumn{7}{c}{\textbf{Model}}\\
\toprule
&&NoReg&\makecell{GNI\\($\sigma$=0.05)}&\makecell{GNI\\($\sigma$=50)}&\makecell{CopyReg\\($\lambda$=0.001)}&\makecell{CopyReg\\($\lambda$=0.01)}&\makecell{SyntaxReg\\($\lambda$=0.001)}&\makecell{SyntaxReg\\($\lambda$=0.01)}\\
\midrule
\multirow{4}{*}{\rotatebox{90}{\textbf{Pre-PT}}}
& Dundee & \textbf{\textcolor{mdred}{-0.971}} & \textbf{\textcolor{mdred}{-0.935}} & \textbf{\textcolor{mdred}{-0.992}} & \textbf{\textcolor{mdred}{-0.975}} & \textbf{\textcolor{mdred}{-0.991}} & \textbf{\textcolor{mdred}{-0.981}} & \textbf{\textcolor{mdred}{-0.975}} \\
 & MECO & \textbf{\textcolor{mdred}{-0.76}} & -0.443 & -0.531 & \textbf{\textcolor{mdred}{-0.821}} & \textbf{\textcolor{mdred}{-0.84}} & \textbf{\textcolor{mdred}{-0.67}} & \textbf{\textcolor{mdred}{-0.892}} \\
 & Provo & \textbf{\textcolor{mdred}{-0.971}} & \textbf{\textcolor{mdred}{-0.767}} & \textbf{\textcolor{mdred}{-0.974}} & \textbf{\textcolor{mdred}{-0.987}} & \textbf{\textcolor{mdred}{-0.979}} & \textbf{\textcolor{mdred}{-0.843}} & \textbf{\textcolor{mdred}{-0.781}} \\
 & NS & \textbf{\textcolor{mdred}{-0.96}} & \textbf{\textcolor{mdred}{-0.894}} & \textbf{\textcolor{mdred}{-0.957}} & \textbf{\textcolor{mdred}{-0.879}} & \textbf{\textcolor{mdred}{-0.905}} & \textbf{\textcolor{mdred}{-0.811}} & \textbf{\textcolor{mdred}{-0.94}} \\
\midrule
\multirow{4}{*}{\rotatebox{90}{\textbf{Post-PT}}} & Dundee & \textbf{\textcolor{mdgreen}{0.811}} & \textbf{\textcolor{mdgreen}{0.519}} & \textbf{\textcolor{mdgreen}{0.513}} & 0.201 & -0.087 & -0.137 & \textbf{\textcolor{mdgreen}{0.793}} \\
 & MECO & 0.044 & \textbf{\textcolor{mdgreen}{0.742}} & \textbf{\textcolor{mdgreen}{0.525}} & -0.32 & -0.261 & -0.151 & \textbf{\textcolor{mdgreen}{0.521}} \\
 & Provo & \textbf{\textcolor{mdgreen}{0.892}} & 0.414 & \textbf{\textcolor{mdgreen}{0.712}} & \textbf{\textcolor{mdgreen}{0.88}} & \textbf{\textcolor{mdgreen}{0.801}} & 0.209 & \textbf{\textcolor{mdgreen}{0.72}} \\
 & NS & 0.294 & 0.132 & \textbf{\textcolor{mdgreen}{0.452}} & -0.221 & \textbf{\textcolor{mdred}{-0.605}} & -0.32 & \textbf{\textcolor{mdred}{-0.737}} \\
 \bottomrule
 \end{tabular}

    \caption{Pearson's $r$ between loss and PPP before (top) and after (bottom) the phase transition. Statistically significant \textbf{\textcolor{mdgreen}{positive}} and \textbf{\textcolor{mdred}{negative}} correlations are boldfaced and colored in \textbf{\textcolor{mdgreen}{green}} and \textbf{\textcolor{mdred}{red}}, respectively.}
    \label{tab:pre_post_corr}
\end{table*}

\section{Loss and PPP}\label{appendix:loss}

As discussed above, comparing PPPs independent of loss is misleading, as they have been found to interact. \Cref{fig:loss} summarize the validation loss of each model throughout the pretraining on 10B tokens. Most models follow a similar pattern, where the loss clearly improves until around 1B tokens, after which the improvement slows down or stops. Copy-regularized model with $\lambda=0.01$ seems to be an exception: validation loss gradually goes \textit{up} from around 1B pretraining tokens, likely due to the strong suppression of induction heads.

\Cref{tab:pre_post_corr} summarizes the loss-PPP correlations before (top) and after (bottom) the phase transition. This corresponds to \Cref{fig:pre_post_corr} in \Cref{sec:exp3:result}. Evidently, most correlations are strongly and significantly negative before the phase transition, whereas about a half of the correlations are significantly positive after the transition. This again highlights the observation that the loss-PPP correlation is indeed flipped (i.e. goes through a tipping point) for both non-regularized and regularized models.

\end{document}